\def\mytitle{Straight-Leg Walking Through Underconstrained Whole-Body Control}
\def\myauthor{Robert J. Griffin\textsuperscript{1,2}, Georg Wiedebach\textsuperscript{2}, Sylvain Bertrand\textsuperscript{2}, Alexander Leonessa\textsuperscript{1}, Jerry Pratt\textsuperscript{2}
\thanks{This work was funded through the NSF NRI Grant No. 1525972 and by the NASA Grant No. NNX12AP97G.}
\thanks{\textsuperscript{1} The author is with the Terrestrial Robotics Engineering \& Controls Lab, Virginia Tech, 635 Prices Fork Rd, Blacksburg, VA 24060, United States }
\thanks{\textsuperscript{2} The author is with the Florida Institute for Human and Machine Cognition, 40 S Alcaniz St, Pensacola, FL 32502, United States}
\thanks{Email : \url{ {rgriffin, sbertrand, gwiedebach, jpratt}@ihmc.us}, \url{{leonessa@vt.edu}}
}}
     
\def\myabstract{abstract}
\def\bibliocommand{\bibliography{mybib}}

\documentclass[10pt,conference,twocolumn]{ieeeconf}
\IEEEoverridecommandlockouts
\usepackage{graphics}
\usepackage{graphicx}
\usepackage{url}
\usepackage{hyperref}
\usepackage{apalike}
\usepackage{tabulary}
\usepackage{booktabs}

\usepackage[
    natbib=true,
    style=ieee,
    url=false,
    doi=false,
    isbn=false,
    safeinputenc=true
]{biblatex}
\AtEveryBibitem{\clearfield{issn}}
\renewcommand{\bibliography}[1]{\addbibresource{#1.bib}}
\bibliocommand

\usepackage{array}
\usepackage{amsmath}
\usepackage{amsfonts}
\usepackage{algorithm}
\usepackage{algpseudocode}

\usepackage{dirtytalk}
\usepackage{xcolor}
\usepackage[english]{babel}

\newcommand{\V}[1] {\boldsymbol{\mathbf{#1}}}

\renewcommand\[{\begin{equation}}
\renewcommand\]{\end{equation}}

\title{\mytitle}

\ifx\myassociation\undefined
\author{\myauthor}
\else
\author{\myauthor \thanks{\myassociation}}
\fi

\begin{document}

\maketitle
\ifx\myabstract\undefined
\else
\begin{abstract}
We present an approach for achieving a natural, efficient gait on bipedal robots using straightened legs and toe-off.
Our algorithm avoids complex height planning by allowing a whole-body controller to determine the straightest possible leg configuration at run-time.
The controller solutions are biased towards a straight leg configuration by projecting leg joint angle objectives into the null-space of the other quadratic program motion objectives.
To allow the legs to remain straight throughout the gait, toe-off was utilized to increase the kinematic reachability of the legs.
The toe-off motion is achieved through underconstraining the foot position, allowing it to emerge naturally.
We applied this approach of under-specifying the motion objectives to the Atlas humanoid, allowing it to walk over a variety of terrain.
We present both experimental and simulation results and discuss performance limitations and potential improvements.
\end{abstract}
\fi

\ifx\mykeywords\undefined
\else
\begin{IEEEkeywords}
\mykeywords
\end{IEEEkeywords}
\fi

\section{Introduction}
\label{introduction}

With few exceptions, nearly all bipedal robot walking gaits utilize highly bent knees, walking with an almost ``squatted'' motion.
Not only is this highly unnatural, it results in significant increase in power consumption at the knee~\citep{Kajita_2003} compared to humans~\citep{Winter_2009}.
Counter to this, passive-dynamic walkers, rely on using only the natural dynamics of walking and have the fundamental characteristic of walking with straight legs~\citep{McGeer_1990}.
This offers energetic benefits, requiring far less torque, and allowing the swing leg to behave like a double pendulum.
Walking with straighter legs also increases the overall ground clearance of the robot, allowing it to step over larger objects and avoid collisions, as well as decreasing the required range of motion and actuator velocities for the same walking speed.
A method for utilizing the natural dynamics of walking to achieve this straight legged gait is needed to allow these benefits to be realized by actively powered robots.

From a control perspective, walking with straight legs poses significant challenges.
First, by allowing height variations, some models such as the Linear Inverted Pendulum (LIP) lose accuracy, as they assume that the center of mass (CoM) is moving through a plane at constant height~\citep{Kajita_2001}.
Second, when the legs of the robot are completely straightened, the ground reaction forces are primarily dependent on the gravity vector and ankle torques, with the control authority from the knee effectively removed.
Last, the straightened leg introduces a singularity in the Jacobian used by inverse-kinematics and inverse-dynamics based approaches.
While there are mechanisms to provide greater control authority, such as step adjustment and angular momentum, and handle Jacobian singularities, the greatest challenge comes from the now-nonlinear dynamics.
These can be solved linearly using a predefined height trajectory, and then controlled using standard zero-moment point (ZMP)~\citep{Morisawa_2005, Kuindersma_2014, Park_2007} or divergent component of motion (DCM)~\citep{Englsberger_2015, Hopkins_2014} approaches.
Also, the height control can be decoupled from the planar control, relying on the robustness of the planar controller.
To walk with straight legs, though, both approaches require complex approaches for computing an appropriate height trajectory.

\begin{figure}[!t]
\centering
\includegraphics[width=2.75in]{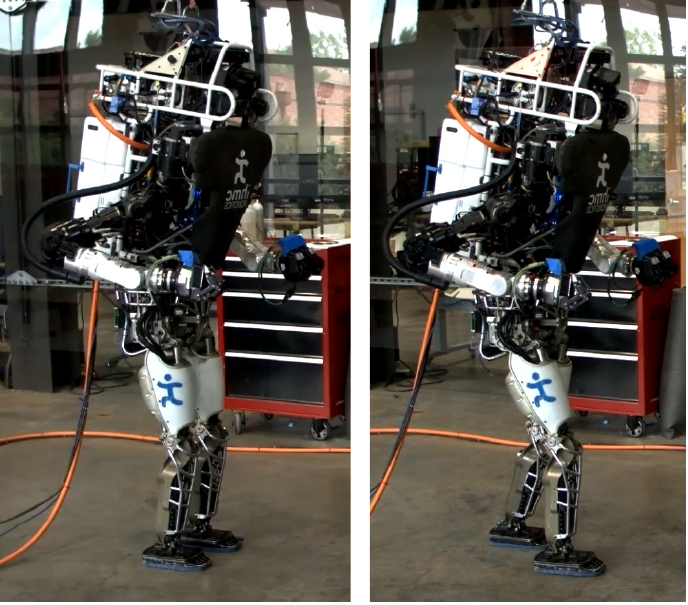}
\vspace{-3mm}
\caption{Images of the Atlas robot standing and stepping with straight legs.}
\vspace{-8mm}
\label{fig:straight_legged_figure}
\end{figure}

Several works have attempted to address this complicated planning problem.
The general objective becomes to design trajectories that result in the robot's legs being as straight as possible without violating the kinematics or the centroidal dynamics.
In~\citep{Li_2010}, the authors presented an approach for generating straightened-knee trajectories by modeling the legs as spring dampers to compute the CoM height after generating horizontal trajectories using model predictive control (MPC) of the ZMP.
Alternatively,~\citep{Brasseur_2015} used linear differential inclusion to incorporate these height trajectories directly into the MPC, resulting in relatively natural, cyclic motions of the CoM.
In~\citep{Heerden_2015}, the authors also included CoM height in their ZMP-based quadratic program, but required sequential quadratic programming to solve the resulting problem's quadratic constraints.
In all cases, after planning, the desired CoM trajectories can be tracked in a momentum-based control framework, as proposed in~\citep{Orin_2013}.

Instead of using complex CoM height planning to walk with straight legs, we propose a new approach that relies the whole-body control framework's, ability to generate torques given a desired set of motion tasks~\citep{Koolen_2016}.
We propose directly address the objective of straightening the legs by biasing the whole-body controller solutions towards those that straighten the legs as much as possible, rather than through setting a desired CoM height.
As noted in~\citep{You_2016}, the whole-body controller can resolve the kinematic and dynamic constraints on the system at run time.
To accomplish this, we propose underconstraining the whole-body controller by not specifying a desired force in the vertical direction.
Instead, we can project a desired straightening leg motion into the null-space of the other objectives.
The robot then will attempt to make its legs as straight as possible for any desired motion, as opposed to trying to achieved a specified CoM height trajectory.
This becomes more similar to the passive-dynamic walking approach, allowing the dynamics of the system to determine the robot height rather than a preplanned trajectory.

Straight leg walking requires several additional elements for proper
execution.
Toe-off is an integral part of natural walking, it increases the number of reachable footholds that do not require significant knee bend by increasing the length of the stance leg.
Toe-off is typically achieved with whole-body controllers by prescribing some toe-off motion in the foot~\citep{Feng_2015, Knabe_2016}.
Instead, we propose to underconstrain the foot motion, as with our approach for straight leg walking.
By under-specifying the desired motions of the foot, the toe-off motion becomes an emergent behavior of the gait.
As the transition to toe-off is state dependent, we define conditions for when the robot should use a toe-off motion.
Additionally, when walking with straight legs, the desired step position is often near the edge of the leg's workspace, with terrain height uncertainties creating significant challenges.
In this work, we propose several mechanisms to compensate for this, as well.

In our approach, we utilize a step adjustment strategy based on that presented in~\citep{Griffin_2017a} to increase the control authority of the robot when walking with straight legs.
We also guarantee the feasibility of the dynamic trajectories for walking with straight legs using the approach presented in~\citep{Griffin_2017b}.
This is executed using the instantaneous capture point (ICP) planner in a whole-body controller framework, as presented in \autoref{controlframework}.
The approach for achieving straight leg walking is described in \autoref{straightlegwalkingcontrol}, with the toe-off approach outlined in \autoref{toe-off}.
Finally, we present both physical and simulation experiments performed both with the Atlas robot in \autoref{resultsanddiscussion}.

\section{Control Framework}
\label{controlframework}

\subsection{Walking Control}
\label{walkingcontrol}

To allow the balance control problem to be tractable for real-time control, reduced order models are typically employed.
The LIP treats the robot as a point mass at the end of a pendulum, whose base is the ZMP~\citep{Kajita_2001}.
The ICP was introduced as an extension of the LIP, and is a state transformation of the CoM defined as
\[
\V{\xi} = \V{x} + \frac{1}{\omega} \dot{\V{x}},
\]
where $\V{x}$ is the center of mass position and $\omega = \sqrt{ g / z_\text{CoM}}$ is the inverted pendulum natural frequency.
The ICP dynamics are then defined as
\[
\dot{\V{\xi}} = \omega \left( \V{\xi} - \V{r}_{\text{CMP}} \right),
\]
where $\V{r}_{\text{CMP}}$ is the Centroidal Moment Pivot~\citep{Popovic_2005}, which is used to control the ICP and encodes the system's ground reaction forces and angular momentum rate.
In this work, we use the ICP planning approach presented in~\citep{Englsberger_2014}.

\begin{figure}[!t]
\centering
\includegraphics[width=2.2in]{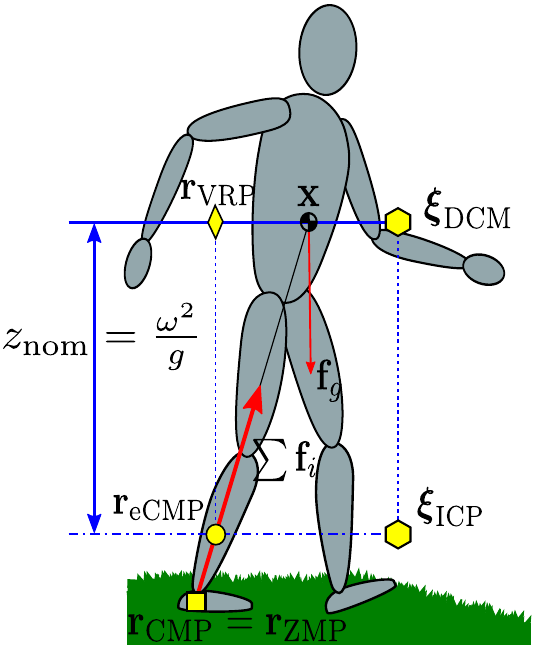}
\vspace{-3mm}
\caption{Diagram showing the relationship between important ground reaction points.
The ICP encodes the CoM position and velocity, and is controlled by the eCMP.
The CMP is lies at the intersection between of the line passing through the CoM and the eCMP and the ground.
For a more detailed definition of the DCM and VRP, see~\citep{Englsberger_2013}}
\label{fig:icp_diagram}
\end{figure}

As it is based on the LIP, the ICP typically assumes that that the CoM moves in a constant height above the ground.
However, $\omega$ can be treated as a design variable, rather than strictly defined by the LIP length, as noted in~\citep{Englsberger_2013}.
The enhanced centroidal moment pivot (eCMP) is located at the intersection of the CMP line with the ICP control plane, which is located $z = \omega^2 / g$ distance below the CoM~\citep{Englsberger_2013}.
The eCMP in this case is used to control the ICP dynamics, rather than the CMP.
The CMP is then typically located at the intersection of the CMP line and the ground plane, as shown in \autoref{fig:icp_diagram}.

\begin{figure}[!t]
\centering
\includegraphics[width=2.9in]{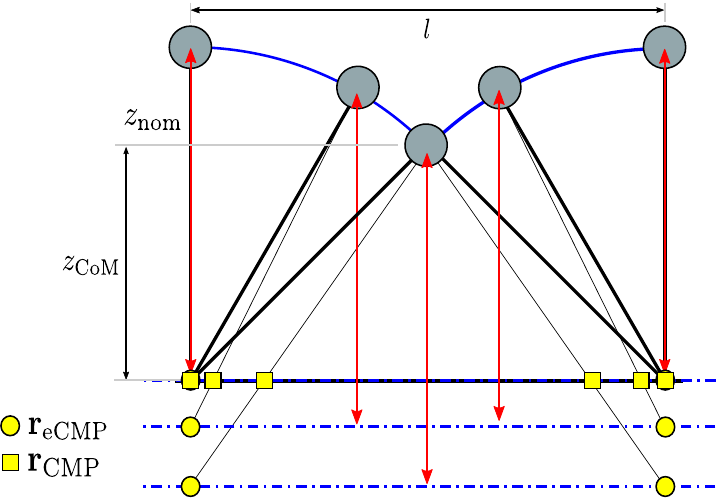}
\vspace{-4mm}
\caption{Diagram showing the CMP error as the pendulum height changes while walking.}
\vspace{-6mm}
\label{fig:walking_deviation}
\end{figure}

As discussed in~\citep{You_2016}, by allowing the CoM height to vary, the CMP deviates from the nominal eCMP location.
From the perspective of the ICP, height variations from the nominal height $z_{\text{nom}} = \omega^2 / g$ will cause the CMP to deviate from the eCMP, with a relationship defined by
\[
\frac{z_{\text{CoM}}}{z_{\text{nom}}} \left( \V{x} - \V{r}_{\text{eCMP}} \right) = \V{x} - \V{r}_{\text{CMP}}.
\label{eqn:cmp_deviation}
\]
Using this relationship, we can perform a simple analysis to determine the maximum amount of deviation to be expected when walking using a simple inverted pendulum model with instantaneous exchanges in support, as shown in \autoref{fig:walking_deviation}.
The maximum CMP error occurs when the CoM is the greatest distance from the desired eCMP, at the point of double support exchange.
For this example, if the stride length is $l$, then this distance is $0.5l$.
Evaluating \autoref{eqn:cmp_deviation}, assuming that the nominal CoM height is the pendulum length, the CMP error is
\[
\begin{aligned}
\V{r}_{\text{CMP}} - \V{r}_{\text{eCMP}} = 0.5 l \left( 1 - \frac{ \sqrt{z_{\text{nom}}^2 - 0.25 l^2}}{z_{\text{nom}}} \right).
\end{aligned}
\]
For approximately human parameters of a pendulum height of 1m and a stride length of 0.75m, this yields a CMP deviation of 2.7cm from the nominal value.
While this is not ideal, a standard ICP feedback controller is easily robust enough to reject the resulting tracking errors with proper planning.

\subsection{Whole-Body Control}
\label{whole-bodycontrol}

Momentum-based control frameworks have been quite effective in implementing compliant, force based control.
Whole-body control represents one of the most common formulations, and gained significant prominence during the DARPA Robotics Challenge.
Almost all in the DRC Finals utilize a quadratic program (QP) to optimize a cost function to resolve multiple motion tasks at every controller time step~\citep{Feng_2016, Knabe_2016, Koolen_2016, Kuindersma_2016}.
The QP formulation found in this work is introduced in~\citep{Koolen_2016}, and is as follows:
\[
\begin{array}{ll}
\min_{\dot{\V{v}}_d, \V{\rho}} & J_{\dot{\V{h}}_d} + J_{\V{J}} + J_{\V{\rho}} + J_{\dot{\V{v}}_d} \\
\text{s.t.} & \V{A}\dot{\V{v}}_d + \dot{\V{A}}\V{v} = \V{W}_g + \V{B}_{\text{CoM}} \V{\rho} + \sum_i \V{W}_{\text{ext},i}, \\
& \V{\rho}_{\text{min}} \le \V{\rho}, \\
& \dot{\V{v}}_{\text{min}} \le \dot{\V{v}}_d \le \dot{\V{v}}_{\text{max}}. 
\end{array}
\]
The terms of the objective function are defined as
\[
\begin{array}{ll}
\text{Momentum Objective:}
& J_{\dot{\V{h}}_d} = \left\| \V{P}_{\dot{\V{h}}} \left( \V{A} \dot{\V{v}}_d - \V{b} \right) \right\|^2, \\ 
\text{Motion Objective:}
& J_{\V{J}} = \left\| \V{P}_{\V{J}} \left( \V{J} \dot{\V{v}}_d - \V{p} \right) \right\|^2, \\
\text{Contact Force Cost:}
& J_{\V{\rho}} = \left\| \V{P}_{\V{\rho}} \V{\rho} \right\|^2, \\
\text{Joint Acceleration Cost:}
& J_{\dot{\V{v}}_d} = \left\| \V{P}_{\dot{\V{v}}_d} \dot{\V{v}}_d \right\|^2,
\end{array}
\nonumber
\]
where:

\begin{itemize}
\item $\dot{\V{v}}_d$ are the desired generalized joint accelerations and $\V{\rho}$ consists of the generalized contact forces~\citep{Koolen_2016}.

\item $\V{A}$ is the centroidal momentum matrix and $\V{b} := \dot{\V{h}}_d - \dot{\V{A}} \V{v}$, where $\dot{\V{h}}_d$ is the desired rate of change of momentum and $\V{v}$ is the joint velocities.

\item $\V{J} = \left[ \begin{array}{ccc} \V{J}_1^T & \dots & \V{J}_k^T \end{array} \right]^T$ and $\V{p} = \left[ \begin{array}{ccc} \V{p}_1^T & \dots & \V{p}_k^T \end{array}\right]^T$ are the concatenated Jacobian matrices and respective objective vectors for each of the $k$ desired motions.

\item $\V{B}_{\text{CoM}}$ is the Jacobian matrix from the generalized contact force frame to the centroidal frame.

\item $\V{W}_{\text{ext},i}$ are the $i$ external wrenches acting on the robot.

\item $\V{W}_{g}$ is the gravitational wrench.

\item $\V{\rho}_{\text{min}} \ge 0$ is the lower bound on $\V{\rho}$, used to enforce contact unilaterality.

\item $\dot{\V{v}}_{\text{min}}$ and $\dot{\V{v}}_{\text{max}}$ are bounds on the joint accelerations, used to enforce joint angle limits.

\item $\V{Q}_{\dot{\V{h}}}, \V{Q}_{\V{J}}, \V{Q}_{\V{\rho}}$, and $\V{Q}_{\dot{\V{v}}_d}$ are positive definite cost function weighting matrices, where $\V{Q}_{(\cdot)} = \V{P}_{(\cdot)}^T \V{P}_{(\cdot)}$ .

\end{itemize}

In this work, we choose to incorporate the linear part of the momentum objective $\dot{\V{h}}_d$ by applying a selection matrix to the momentum objective,
\[
J_{\dot{\V{h}}_d} = \left\| \V{P}_{\dot{\V{h}}} \V{S} \left( \V{A}\dot{\V{v}}_d - \V{b} \right) \right\|^2,
\]
where
\[
\V{S} = 
\left[ \begin{array}{cccccc}
0 & 0 & 0 & 1 & 0 & 0 \\
0 & 0 & 0 & 0 & 1 & 0 \\
0 & 0 & 0 & 0 & 0 & 1
\end{array} \right],
\label{eqn:linear_momentum_selection_matrix}
\]
selecting only the linear momentum rate of change, $\dot{\V{l}}$, from $\dot{\V{h}} = \left[ \dot{\V{k}}^T \ \dot{\V{l}}^T \right]^T$.
This leaves the angular momentum rate of change $\dot{\V{k}}$ unconstrained, free to be determined by the optimization.
This allows angular momentum to be generated by natural motions, such as the swing foot, without requiring it be canceled via other motions.

\section{Straight Leg Walking Control}
\label{straightlegwalkingcontrol}

\begin{figure}[!t]
\centering
\includegraphics[width=3.4in]{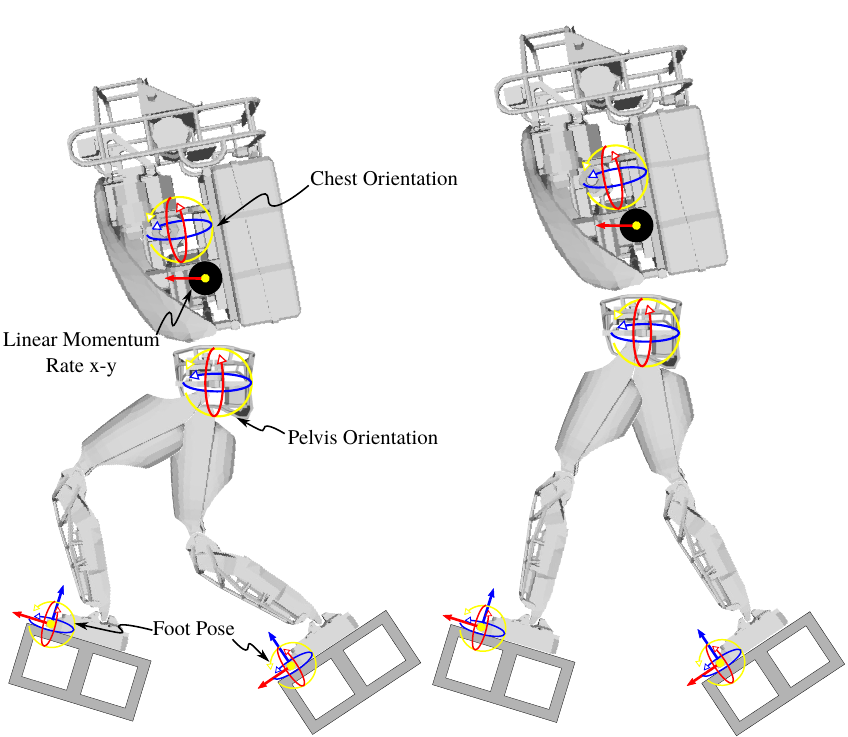}
\vspace{-8mm}
\caption{When the vertical momentum rate of change is not specified, there are many potential solution motions contained in the overall task Jacobian.
This shows two possible configurations, one with bent legs, one with with straight legs.}
\vspace{-6mm}
\label{fig:different_height_solutions}
\end{figure}

Instead of controlling the CoM height directly, we propose controlling it inside the task null-space using desired leg configurations.
The general objective of advanced CoM height planning is to keep the legs as near straight as possible to avoid higher torques at the knee.
Instead of trying to straighten the legs through complicated planning, we instead bias the whole-body controller solution towards a desired configuration.
To do so, we modify the selection matrix in \autoref{eqn:linear_momentum_selection_matrix} to use only the $x$-$y$ momentum rate of change,
\[
\V{S} = 
\left[ \begin{array}{cccccc}
0 & 0 & 0 & 1 & 0 & 0 \\
0 & 0 & 0 & 0 & 1 & 0
\end{array} \right],
\label{eqn:horizontal_momentum_selection_matrix}
\]
leaving the vertical force of the robot unconstrained.
This works well within the instantaneous capture point (ICP) based framework used in~\citep{Koolen_2016}, which required a separate controller on either the CoM or pelvis height for the vertical momentum rate of change.
As typical motion inputs for $\V{p}$ consist of tasks such as pelvis angular acceleration, chest angular acceleration, arm joint acceleration, and swing foot linear and angular acceleration, the resulting QP has a large null-space, with many potential solutions satisfying these objectives, as illustrated in \autoref{fig:different_height_solutions}.

To control the leg configuration without violating any of the higher level motion tasks, we can project desired joint angles, which we will refer to as the \emph{privileged configuration}, into the null-space of the QP.
The privileged configuration, $\V{q}_p$, can be then used to compute the privileged joint accelerations using a simple feedback control law
\[
\dot{\V{v}}_p = \V{k}_p \left( \V{q}_p - \V{q} \right) + \V{k}_d \V{v},
\]
where $\V{q}$ is the current joint position.
These privileged accelerations can then be projected into the null-space of the QP as an additional quadratic motion objective
\[
J_p = \left\| \left( \V{I} - \V{J}_{\text{task}}^{+} \V{J}_{\text{task}} \right) \dot{\V{v}}_p \right\|_{\V{Q}_p}^2,
\]
where $\V{J}_{\text{task}} = \left[ \begin{array}{cc} 
\left(\V{SA}\right)^T & \V{J}^T \end{array} \right]^T$ is the total task Jacobian that maps all the desired inputs to the QP to joint acceleration space, and $\left( \cdot \right)^{+}$ is the pseudo-inverse operator.

This approach of applying privileged configurations to bias QP solutions has been well used to help with singularity escape.
The effect is that, when a joint is at a singularity, there is a non-empty null-space, allowing the privileged configuration to provide an acceleration in the correct direction.
Instead of using it as a mechanism to help with singularity escape, however, we use it to bias the general robot behavior, encouraging the solutions to always use as straight of knees in the support leg as possible.

\subsection{Leg Configuration Selection}
\label{legconfigurationselection}

\begin{figure}[!t]
\centering
\includegraphics[width=3.4in]{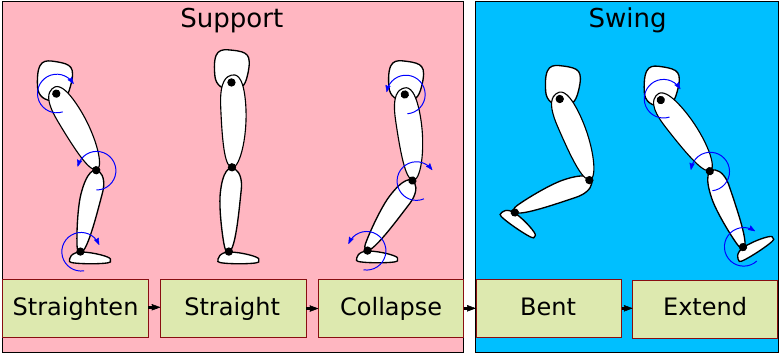}
\vspace{-6mm}
\caption{State machine for the desired leg configuration angle, which straightens and collapses the leg based on the phase of the walking gait.
}
\vspace{-6mm}
\label{fig:leg_state_machine}
\end{figure}

To effectively control the leg configuration, we introduce a state machine that sets the privileged configuration for each leg, shown in \autoref{fig:leg_state_machine}.
This is used to allow the leg to bend at the desired phases in the gait cycle.
On touchdown, the new support transitions to the \say{Straighten} state, where the privileged configuration transitions to the straight configuration over a period of defined time.
At this point, the leg is automatically transitioned into the \say{Straight} state.
Once the support leg is partway through the swing phase, the configuration is changed to \say{Collapsed}, allowing the leg to slightly bend.
This allows the swing foot to track by moving the workspace and assists with toe-off.
The trailing leg then remains in the collapsed state during transfer.
At the beginning of swing, the leg is switched to the \say{Bent} state which helps pick the foot up off the ground and escape the knee singularity.
Partway during the swing phase, it is then switched to the \say{Extend} state, which extends the leg outward in preparation for touchdown.
On touchdown, the leg is then transitioned back to the \say{Straighten} state.

\subsection{Improved Swing Foot Touchdown}
\label{improvedswingfoottouchdown}

When walking, it is generally assumed that the swing foot will strike the ground at the correct time.
While the upcoming foothold is almost always in the swing leg's workspace when the knees are heavily bent, this may not be the case when the stance leg is straight.
If the ground plane is lower than the predicted footstep, the foot will not have come into contact when planned, and will most likely be at the edge of the workspace
This result in an exponentially increasing ICP error with respect to time.
To maintain stable walking, then, the foot must be set down as quickly as possible after its expected touchdown to make the gait robust to height uncertainties.
This is similar to starting to shift weight onto the swing leg, whether it's in contact or not, trusting that it will hit the ground.

A standard approach for handling this is to command a constant downward velocity at the end of the swing foot trajectory.
In addition, we linearly increase the weight on the swing foot spatial accelerations in the whole-body controller with respect to the time past the expected swing touchdown.
This increases the importance of continuing the downward motion of the foot.
We also decrease the weight on the pelvis orientation with respect to this exceeded time, allowing the robot to rotate the pelvis to achieve the desired swing foot motion, similar to how people rotate their pelvis to increase their leg's workspace.

\section{Toe-Off}
\label{toe-off}

Toe-off is an essential part of human-like bipedal walking.
Besides the metabolic savings it provides through capitalizing on the leg's muscle-tendon for energy storage, it is kinematically beneficial, allowing the trailing leg to be extended and the leading leg to bend less.
Toe-off also helps address range of motion limitations, such as when stepping down.
However, when to toe-off is unclear, as decreasing the support polygon size limits the control authority of the robot, as shown in \autoref{fig:toe_off_criteria}.
Additionally, how to achieve toe-off poses many challenges.

\subsection{Toe-Off Criteria}
\label{toe-offcriteria}

\begin{figure}[!t]
\centering
\includegraphics[width=3.32in]{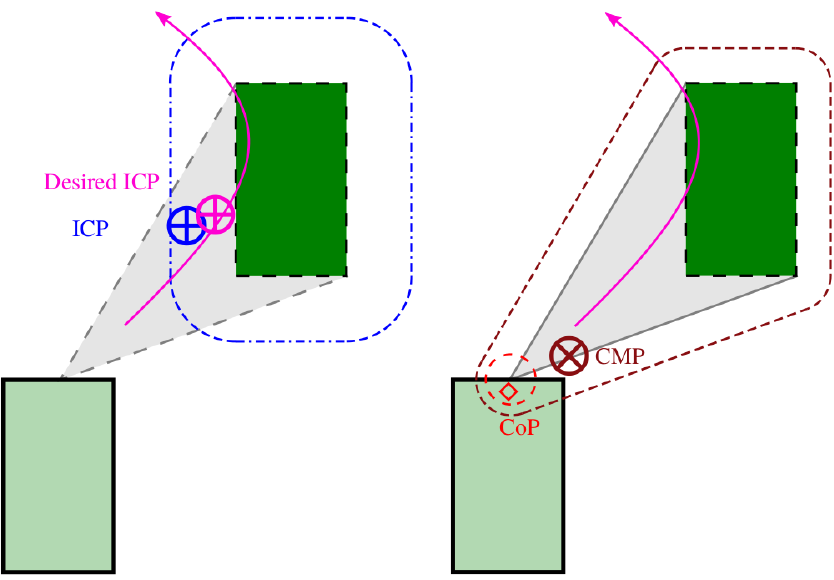}
\caption{Criteria checks for toe-off.
To start toe-off, both the desired ICP and estimated ICP locations must be inside the predicted support polygon (gray), as well as within a certain proximity of the desired foothold (the dotted blue line).
The trailing foot CoP must be less than a certain distance from the toe-off point (the dotted red line), and the desired CMP location must be close to the predicted support polygon, the dark dotted red line.}
\label{fig:toe_off_criteria}
\end{figure}

\begin{figure}[!t]
\centering
\includegraphics[width=3.25in]{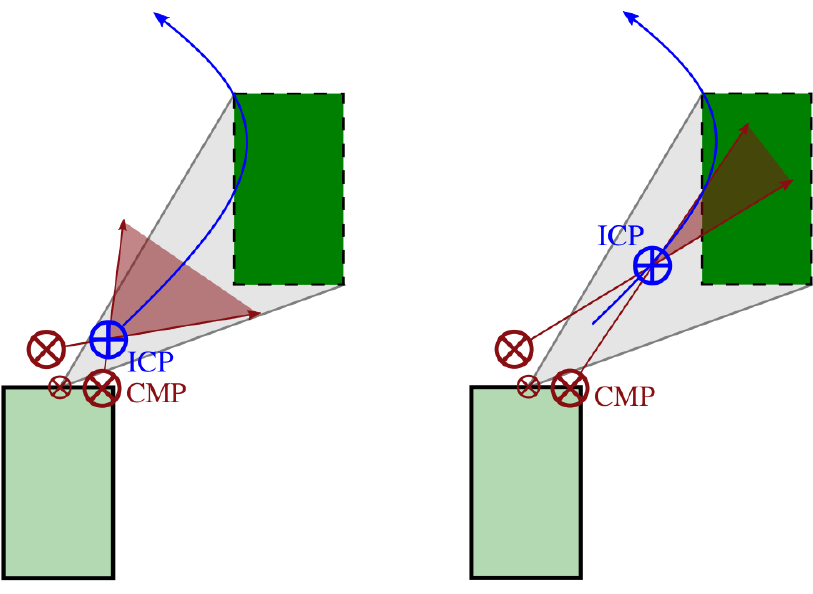}
\caption{Figure showing the effects of ICP location on ICP control for proximity to upcoming foot.
The closer the ICP location is to the stance foot, the more susceptible the ICP dynamics are to CMP location errors, as shown on the left.
If the ICP is closer to the desired foothold, CMP errors affect the direction of the ICP dynamics less, going more in the direction of the foothold.}
\label{fig:toe_off_effects}
\end{figure}

To determine when to transition to toe-off, we establish a set of criteria that must be satisfied.
The support polygon in toe-off is defined using a single toe-off point in the foot, as shown in \autoref{fig:toe_off_criteria}.
If moving to toe-off results in either the desired or current ICP being outside the support polygon, then we no longer have full controllability of the ICP.
This is equivalent to saying that the robot must be 1-step capturable during swing and 0-step capturable during transfer (including toe-off).
The desired and current ICP must also be within a certain distance of the foothold.
Due to torque limits and other uncertainties, the desired CMP may not be achieved.
By increasing the distance of the ICP from the CMP, the effects of the inaccuracies are decreased, as shown in \autoref{fig:toe_off_effects}, and are more likely to move in the direction of the foothold.

Additional toe-off criteria are related to the ground reaction forces.
As changing the support polygon changes the constraints on the center of pressure (CoP), we require that the CoP in the support\slash trailing foot be within a certain proximity of the toe-off polygon, minimizing the change on toe-off.
To minimize the amount of angular momentum, the distance from the centroidal momentum pivot (CMP) to the support polygon must also be minimized.
These requirements are outlined in \autoref{fig:toe_off_criteria}.

\subsection{Toe-Off Control}
\label{toe-offcontrol}

\begin{figure}[!t]
\centering
\includegraphics[width=3.2in]{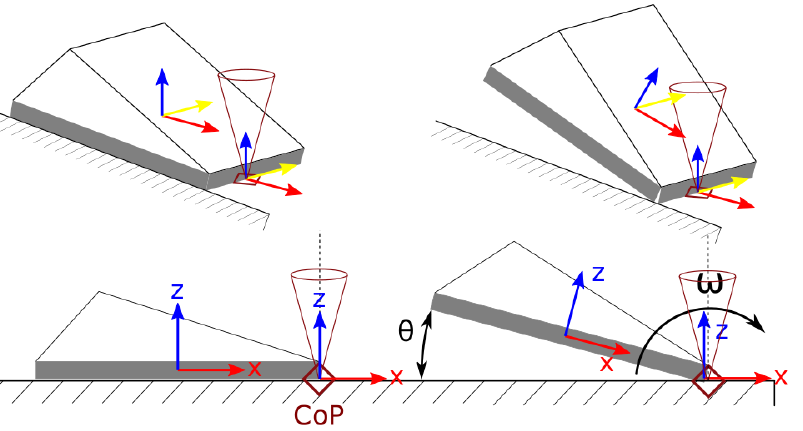}
\caption{During toe-off, a single contact point is enabled on the front of the trailing foot.
The pitch of the foot is then left uncontrolled, while the contact point is held constant in the world.}
\label{fig:toe_off_control}
\end{figure}

Unlike other approaches~\citep{Feng_2015, Knabe_2016}, to control the foot during toe-off, we provide no direct control of the toe-off motion, that is, the pitching of the foot.
Instead we leave this degree of freedom free and unspecified.
This allows the controller to use the foot pitch to achieve other motion tasks, making toe-off motion an emergent behavior.
For the single foot contact point in toe-off, we use the standard friction cone constraints, as shown in \autoref{fig:toe_off_control}.
We found that using a single point resulted in more stable behavior than a contact line.
To perform this control, we can break it into two, separate parts: position and orientation.

During toe-off, we do not want the toe-off contact point of the foot to slip.
This can be achieved using a simple feedback controller to find linear accelerations to hold the point constant in the world frame.
An orientation controller in the sole-frame of the foot can be used to maintain a constant roll and yaw of the foot throughout the toe-off motion.
This leaves the pitch of the foot unconstrained and open in the null-space.
Pitching of the foot about the toe-off point then occurs naturally during the toe-off state, as shown in \autoref{fig:toe_off_control}.

\section{Results and Discussion}
\label{resultsanddiscussion}

\begin{figure*}[!t]
\centering
\includegraphics[width=7.0in]{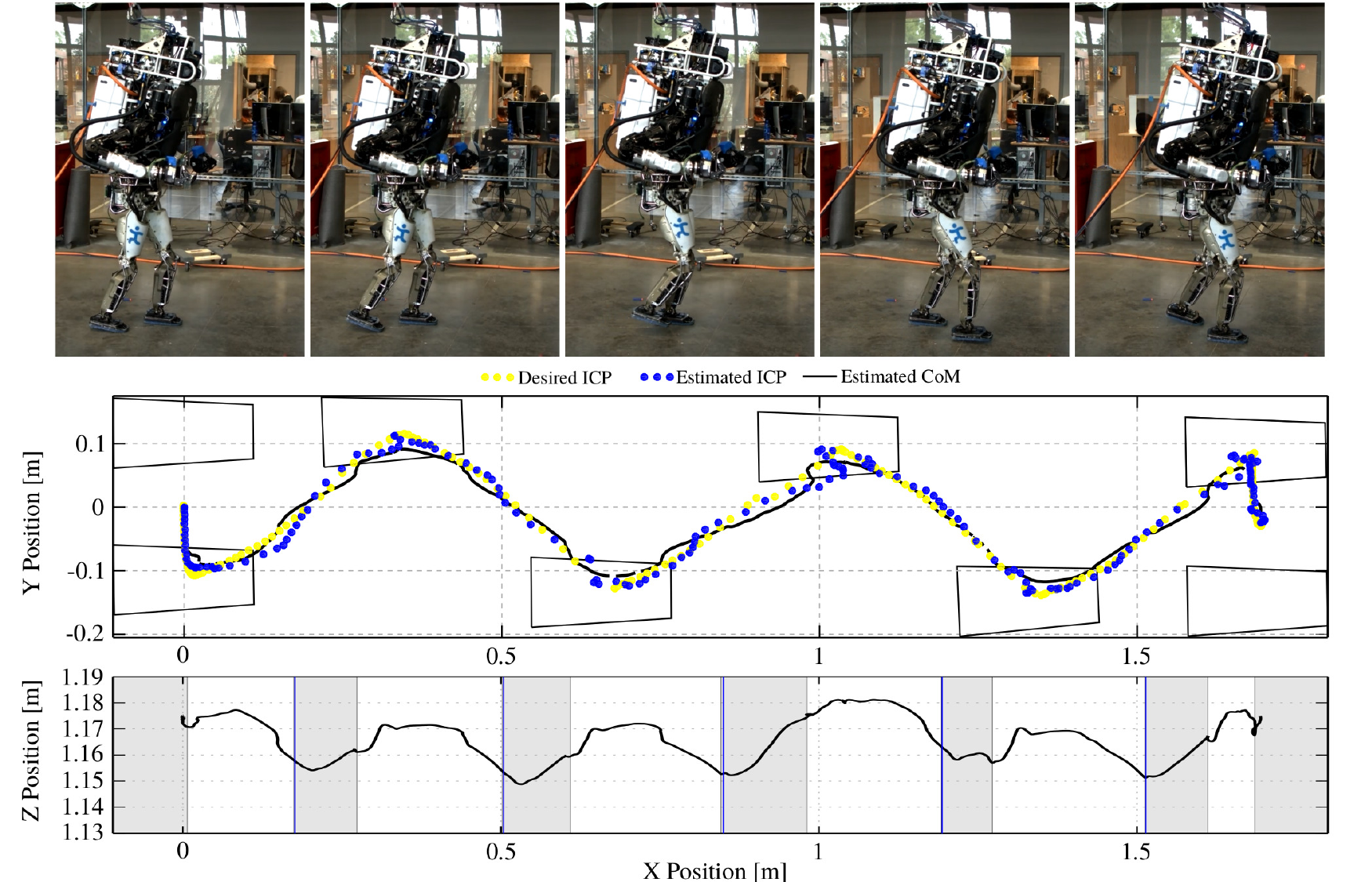}
\vspace{-4mm}
\caption{Atlas walking over flat ground using the proposed method for indirectly controlling the CoM height and toe-off.
The top show several images of a single step, while the bottom figures show the trajectories executed while walking.
The ICP is able to be tracked fairly well using the ICP control framework in \cite{Griffin_2017a}. 
The CoM height trajectory follows a path similar to that of the inverted pendulum, with velocity direction changes occurring in the transfer phase, shown in gray.
Toe-off greatly improved this motion, and is indicated by the blue vertical line.}
\label{fig:flat_ground_walking}
\end{figure*}

We implemented the proposed method on the Atlas robot, and conducted both simulations and physical experiments.
Using the combination of step adjustment from~\citep{Griffin_2017a}, step time optimization from~\citep{Griffin_2017b}, and the presented strategies for straight leg walking and toe-off control, the robot was able to walk over both flat ground and varying terrain with straight legs.
For the physical experiments presented here, a swing time of 1.0s and transfer time of 0.35s was used, while simulation used 0.60s and 0.25s, respectively.
The speeds on hardware are limited by the pump curve of the hydraulic pump, leading to step durations nearly twice that of humans, resulting in less dynamic ICP plans.
This, in turn, results in the robot requiring slightly more bent knees than otherwise possible.

\autoref{fig:flat_ground_walking} shows the resulting ICP and CoM trajectories for Atlas taking 0.35m steps over flat ground, for a walking speed of 0.26m\slash s.
As can be seen, the robot tracks the desired ICP trajectory, albeit it with some error, resulting in a sinusoid-like CoM motion in the $x$-$y$ plane, similar to that of human walking.
By commanding the legs to straighten, the CoM height follows a path similar to that of the inverted pendulum.
Some error occurs when the leading leg is allowed to straighten and overshoots at the beginning of swing.
Unlike the true inverted pendulum, which has sharp changes in height as seen in \autoref{fig:walking_deviation}, the robot's CoM smoothly changes height during the transfer phase, shown in gray in \autoref{fig:flat_ground_walking}.
Toe-off, which is indicated by the vertical blue line, is required for this exchange to occur during transfer.
Without toe-off, the CoM height could not increase until swing, as it allows the trailing leg to extend further, while the leading leg compresses.
By allowing toe-off and height variations, the resulting CoM height trajectory then resembles that observed in humans~\citep{Kuo_2005}.

\begin{figure}[!t]
\centering
\includegraphics[width=3.4in]{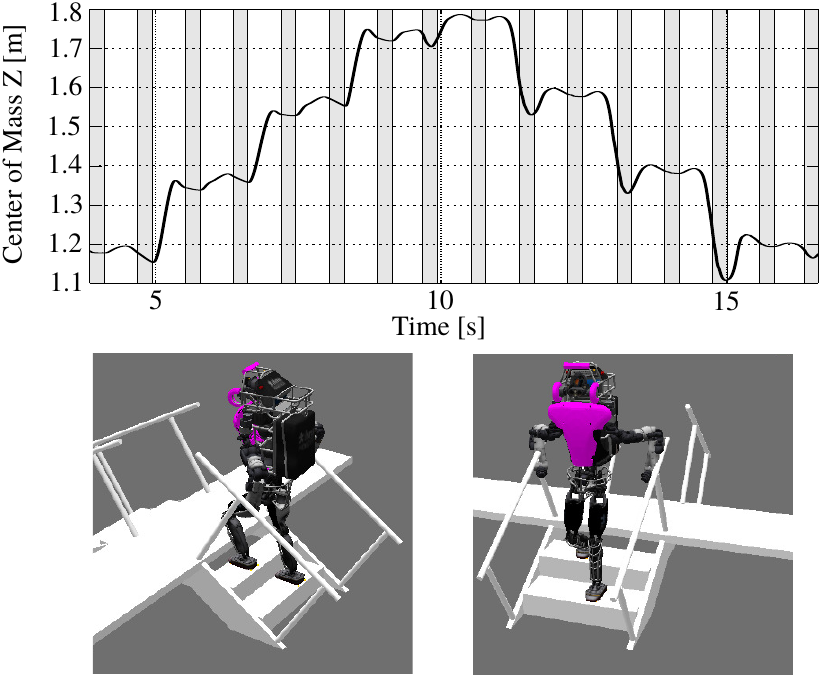}
\caption{Plot of the robot ascending and descending stairs with straight legs in simulation. The stair height is 19.685 cm, the maximum allowable stair height in the US.}
\label{fig:stairs_plot}
\end{figure}

Using the proposed approach, the robot was capable of walking over terrain of varying height.
A simulation of Atlas walking up and down a set of stairs is shown in \autoref{fig:stairs_plot}, with each step 19.685cm (the maximum height allowed by the American's with Disabilities Act) high.
While ascending the stairs posed little challenge for the controller, descending was slightly harder, as the desired swing foot position was often at the edge of the leg's workspace.
To appropriately get the foot down to the next step, the robot had to collapse the stance leg quickly, essentially dropping onto the upcoming step, similar to human.
This was greatly assisted by our proposed approach in \autoref{improvedswingfoottouchdown}, which incentivised the controller to collapse the stance leg and rotate the pelvis to track the desired swing foot position.
The cyclic height trajectories emerged naturally, with the robot raising the CoM during transfer as much as possible, and then stepping up.

\begin{figure*}[!t]
\includegraphics[width=7in]{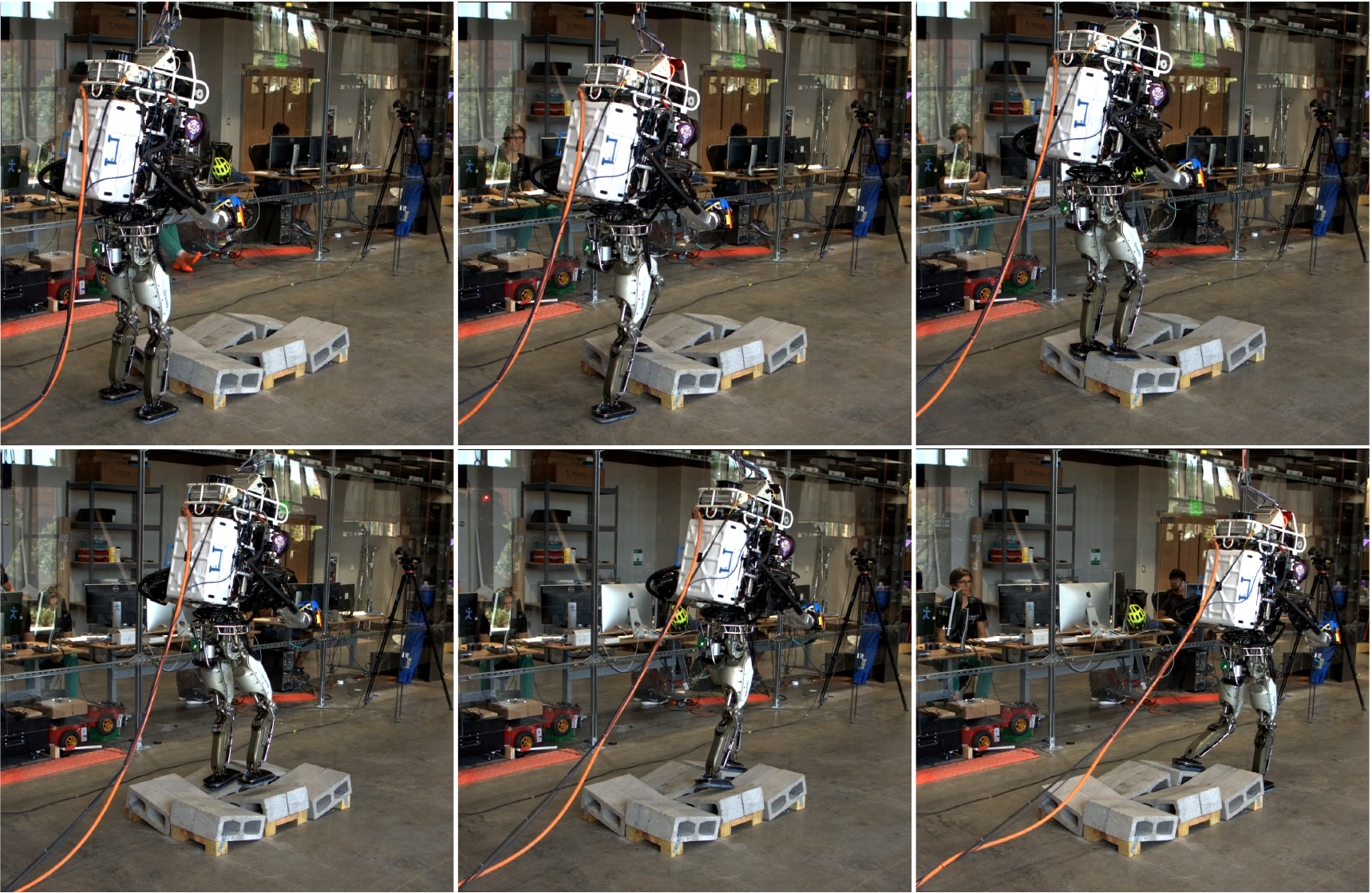}
\caption{Atlas walking over cinder blocks with an indirectly controlled CoM height}
\label{fig:cinder_block_walking_images}
\end{figure*}

The proposed algorithm also was shown to work well on a variety of terrain during hardware experiments.
As shown in \autoref{fig:cinder_block_walking_images}, Atlas was able to walk over cinder blocks of differing heights.
We plan to further test the algorithm over other terrains and stairs to improve tuning, control, and transition criteria.

While the robot has not demonstrated the same level of balance as when walking with bent knees, this is expected, as the control authority is diminished through greatly reducing the effective actuation at the knee.
Additionally, our balance algorithms require adequate control of the CoP and CMP locations, and accurate measurements of both the ICP position and velocity.
While we are normally able to control the CoP to approximately 2cm accuracy, the control accuracy is reduced as the legs are straightened.
As such, we set the desired knee angle to 0.3rad, which we found to be the minimum angle at which we had adequate CoP control for walking.
The CoP inaccuracies start to impose limitations as the desired CMP is moved near the edge of the support polygon.
This can then cause tipping about the edge of the foot, which further exacerbates state estimation problems faced in the ICP calculation.
We plan on exploring improving this by limiting the CoP location to lie a certain distance away from the edge of the support polygon, and relying on angular momentum to achieve CMP locations near or past these bounds.
As actuator torque and velocity limits are improved, the allowable angular momentum rate and stepping speed will increase, providing much more control authority when stepping.
Additional instabilities when walking occur due to stiction in the ankle actuators.
The rapid switching from high-impedance control at the ankles during swing to low-impedance control during support has stability issues when dealing with the higher impulse at heel-strike when walking with straight legs.
This then requires significant damping in the leg control for the walking to remain stable, which will be improved with better force control.

\section{Conclusion}
\label{conclusion}

Using the presented approach, the Atlas humanoid was capable of walking with straight legs over a variety of terrains both in simulation and on hardware.
This was made possible by a novel approach that under-specified the desired motion tasks for the robot.
This includes indirectly controlling the center of mass height by biasing the whole-body controller towards solutions that use straight legs, and allowing the whole-body controller to perform toe-off motions when necessary through leaving foot pitch uncontrolled.
We believe this has several important implications.
First, more efficient, natural-looking walking gaits do not require complicated planning, and can be achieved by leaving the height uncontrolled, allowing the whole-body controller to determine when and when not to use straight legs.
Second, many motions of the robot may be over-controlled, with natural behaviors emerging when some of the objectives are removed.
The whole-body controller can then determine what motions are necessary to achieve the essential tasks, such as desired balancing forces.

In the future, we hope to further the robustness of this approach to height uncertainties, enabling the robot to traverse similar terrains as humans.
This includes incorporation of our work in~\citep{Wiedebach_2016}, and improving step adjustment strategies to work on rough terrain using convex polygon constraints.
We also plan on exploring new strategies for controlling the stance leg configuration, increasing the robustness of the controller to terrain uncertainties.


\ifx\myappendix\undefined
\else

\section*{APPENDIX}
\input{\myappendix}

\fi

\ifx\myacknowledgments\undefined
\else

\section*{ACKNOWLEDGMENT}
\input{\myacknowledgments}

\fi

\ifx\bibliocommand\undefined
\else
\printbibliography
\fi

\end{document}